\documentclass{article}
\usepackage{graphicx}
\usepackage{natbib}
\usepackage{authblk}
\usepackage{amsmath}
\usepackage{amssymb}
\usepackage{bbm}
\usepackage{subcaption}
\usepackage{algorithm}
\usepackage{algpseudocode}
\usepackage{multirow}
\DeclareMathOperator*{\argmax}{argmax}
\DeclareMathOperator*{\E}{\mathbb{E}}
\DeclareMathOperator{\R}{\mathbb{R}}
\DeclareMathOperator{\N}{\mathbb{N}}
\DeclareMathOperator{\ind}{\mathbbm{1}}

\algrenewcommand\algorithmicrequire{\textbf{Input:}}
\algrenewcommand\algorithmicensure{\textbf{Output:}}
\begin{document}

\title{Active Deep Q-learning with Demonstration
}

\author[1,2]{Si-An Chen\thanks{r05922089@ntu.edu.tw}}
\author[2]{Voot Tangkaratt\thanks{voot.tangkaratt@riken.jp}}
\author[1]{Hsuan-Tien Lin\thanks{htlin@csie.ntu.edu.tw}}
\author[2,3]{Masashi Sugiyama\thanks{sugi@k.u-tokyo.ac.jp}}

\affil[1]{National Taiwan University, Taiwan}
\affil[2]{RIKEN Center for Advanced Intelligence Project, Japan}
\affil[3]{The University of Tokyo, Japan}


\date{}

\maketitle

\begin{abstract}
Recent research has shown that although Reinforcement Learning (RL) can benefit from expert demonstration, it usually takes considerable efforts to obtain enough demonstration. The efforts prevent training decent RL agents with expert demonstration in practice. In this work, we propose Active Reinforcement Learning with Demonstration (ARLD), a new framework to streamline RL in terms of demonstration efforts by allowing the RL agent to query for demonstration actively during training. Under the framework, we propose Active Deep Q-Network, a novel query strategy which adapts to the dynamically-changing distributions during the RL training process by estimating the uncertainty of recent states. The expert demonstration data within Active DQN are then utilized by optimizing supervised max-margin loss in addition to temporal difference loss within usual DQN training. We propose two methods of estimating the uncertainty based on two state-of-the-art DQN models, namely the divergence of bootstrapped DQN and the variance of noisy DQN. The empirical results validate that both methods not only learn faster than other passive expert demonstration methods with the same amount of demonstration and but also reach super-expert level of performance across four different tasks.
\end{abstract}

\section{Introduction}
\label{intro}
Sequential decision making is a common and important problem in the real world. For instance, to achieve its goal, a robot should produce a sequence of decisions or movements according to its observations over time. A recommender system should decide when and which item or advertisement to display to a customer in a sequential manner. For sequential decision making, reinforcement learning~\citep{DBLP:books/lib/SuttonB98} (RL) has been recognized as an effective framework which learns from interaction with the environment. Thanks to advances in deep learning and computational hardware, deep RL has achieved a number of successes in various fields such as end-to-end policy search for motor control~\citep{DBLP:conf/nips/WatterSBR15}, deep Q-networks for playing Atari games~\citep{DBLP:journals/nature/MnihKSRVBGRFOPB15}, and combining RL and tree search to defeat the top human Go expert~\citep{DBLP:journals/nature/SilverHMGSDSAPL16}. These successes show the power of RL to solve various kinds of sequential decision making and control problems.

In contrast with these successes, deep RL algorithms are notorious for their substantial demands on simulation during training. Typically, these algorithms start from scratch and require millions of data samples to learn a locally optimal policy, which is not a problem if unlimited simulation is available but is infeasible for many real-world applications such as robotic systems. To address this problem, several methods have been proposed to improve learning efficiency by leveraging prior knowledge from human experts. Imitation learning~\citep{DBLP:conf/nips/Schaal96}, also known as learning from demonstration (LfD), is an attempt to learn the policy of an expert by observing the expert's demonstrations. 
However, the performance of imitation learning is limited by the expert, since the agent only learns from the expert without regard to rewards given by the environment. Another way is to improve RL by leveraging demonstrations given by the expert and rewards simultaneously. Recently, Deep Q-learning from Demonstration (DQfD)~\citep{DBLP:conf/aaai/HesterVPLSPHQSO18} and Policy Optimization with Demonstrations (POfD)~\citep{pmlr-v80-kang18a} have shown state-of-the-art results on several Atari games by training the agent with an objective that combines the rewards and the expert demonstrations.

Although expert demonstrations improve RL, the efforts made by the expert are not negligible. For instance, it takes a human expert thousands of steps to finish a mere 5 episodes for most Atari games~\citep{DBLP:conf/aaai/HesterVPLSPHQSO18}. The huge efforts make it hard to collect a large number of demonstrations for DqFD in practice. In this paper, we introduce the concept of active learning to make more efficient use of the expert's efforts. In supervised learning, the goal of active learning is to achieve better performance with less labeling effort by interactively querying for new labels from unlabeled data~\citep{settles.tr09}. In RL, we can also actively ask the expert for a recommended action given the current observed state. Videos of bootstrapped DQN~\citep{DBLP:conf/nips/OsbandBPR16} have shown that the behavior of different well-performing policies agree at critical points but diverge at other less important states. This suggests that we could save much expert effort by querying only at critical states in contrast to previous methods in which the expert's demonstration is collected for several entire episodes. In other words, we can achieve further improvement in RL with the same number of demonstrations.

In this work, we consider reinforcement learning problems that allow \textit{selective} human demonstration \textit{during} the learning process. We first propose a new framework called Active Reinforcement Learning with Demonstration (ARLD) for such learning problems.
Then we propose Active DQN, which proactively asks for demonstration and leverages the demonstration data. The query criterion should decide when to query---i.e., identifying states where the agent can indeed learn and improve by obtaining the demonstration. We propose two query methods based on uncertainty of Q-value estimation, named divergence and variance, which are derived from two state-of-the-art DQN methods, bootstrapped DQN and noisy DQN, respectively. 
The uncertainty terms are then thresholded dynamically to form querying decisions. 

The dynamic nature allows the two methods to adapt to recent states observed by the agent and can be applied in various kinds of environments without exhaustive parameter tuning. Experimental results show that our method with both uncertainty measurements is effective in four different tasks. In this paper we thus offer three main contributions:
\begin{enumerate}
	\item We propose a new framework, Active Reinforcement Learning with Demonstration, which is the first work to reduce human effort in RL with demonstration to the best of our knowledge;
	\item We propose a novel uncertainty-based query strategy which can be applied toward different tasks and less sensitive to additional parameters;
    \item We verify the effectiveness of two DQN uncertainty estimations with promising experiment results.
\end{enumerate}

\section{Related Work}
\label{sec:rel}

Imitation learning~\citep{DBLP:conf/nips/Schaal96} is a classic technique for learning from human demonstration. Typically, imitation learning uses a supervised approach to imitate an expert's behaviors. DAGGER~\citep{DBLP:journals/jmlr/RossGB11}, a popular imitation algorithm, requests an action from the expert at each step, and takes an action sampled from a mixed distribution of the agent and the expert. It then aggregates the observed states and demonstrated actions to train the agent iteratively. Deeply AggreVaTeD~\citep{DBLP:conf/icml/SunVGBB17} is an extended version of DAGGER which works with deep neural networks and continuous action spaces. However, both require an expert to provide demonstration during the whole training phase. To reduce the demand for human effort, the agent learns actively in active imitation learning~\citep{DBLP:conf/aaai/ShonVR07,judah2014active}
by requesting fewer expert demonstrations. The supervised setting of imitation learning make it easier to apply techniques from traditional supervised active learning. However, although imitation learning can lead to no-regret performance in online learning settings, its performance is still limited by the expert given the use of \textit{only} expert demonstration data for learning. 

On the contrary, it is possible for Reinforcement Learning (RL) to achieve better performance than the human expert by learning to interact with the environment and maximize the cumulative rewards. In RL, there exist a variety of methods that leverage demonstration to obtain improved performance. For instance, some use expert advice or demonstration to shape rewards in the RL problem~\citep{DBLP:conf/ijcai/BrysHSCTN15,DBLP:conf/atal/SuayBTC16}. Another approach is to ask for demonstration from a given state to another state to improve the exploration~\citep{DBLP:conf/atal/SubramanianIT16}. In contrast, the HAT algorithm summarizes the demonstrated knowledge via a decision tree and bootstraps the task with the learned policy to transfer it to the target agent~\citep{DBLP:conf/atal/TaylorSC11}. CHAT, an extension of HAT, measures the source policy's confidence to decide whether to take its advice~\citep{DBLP:conf/ijcai/WangT17}. The main difference between CHAT and our work is CHAT's use of another model to learn a source policy from demonstration offline and estimate the confidence of the source policy; in contrast we estimate the uncertainty of the target learner and ask the expert directly.

Reinforcement Learning with Expert Demonstrations (RLED)~\citep{DBLP:conf/pkdd/PiotGP14} concerns a scenario in which the expert also receives rewards from the environment. In this case, DQfD~\citep{DBLP:conf/aaai/HesterVPLSPHQSO18}, DDPGfD~\citep{DBLP:journals/corr/VecerikHSWPPHRL17}, and POfD~\citep{pmlr-v80-kang18a} have shown state-of-the-art results on a variety of tasks by combining the original RL loss with a supervised loss on the expert's demonstrations. Then, the agent simultaneously learns its original objective and the behavior of the expert. In comparison to similar work such as Human Experience Replay~\cite{DBLP:journals/corr/HosuR16} and Replay Buffer Spiking~\cite{DBLP:journals/corr/LiptonGLLAD16}, RLED methods yield massive acceleration with a relatively small amount of demonstration data. Moreover, experiments show that these methods can also outperform the expert they learn from. In contrast to our work, these works collect demonstration data before training, and the expert must interact with environment by completing the whole episode several times,
whereas the proposed method requires the expert to demonstrate only at critical states given the learning progress of the agent.

Most previous works focus on how to improve RL from ``passive'' demonstration data. To the best of our knowledge, this is the first work to introduce the concept of active learning to leverage demonstration data. The most similar work is active imitation learning; however, as the mechanisms for supervised learning and reinforcement learning differ greatly, we cannot directly apply the same methods. Table \ref{t:setting} compares different settings mentioned above.

\begin{table}[ht]
\resizebox{\linewidth}{!}{
\begin{tabular}{l|c|c|c}
                   & no expert & offline/passive learning & online/active learning\\
\hline
\multirow{2}{6em}{Imitation Learning}    &   & DAGGER & \multirow{2}{8em}{Active Imitation Learning}\\
 & & Deeply AggreVaTeD & \\
\hline
\multirow{4}{6em}{Reinforcement Learning} & DQN & DQfD & \multirow{4}{8em}{\bf{ARLD \newline (our work)}} \\
 & DDPG & DDPGfD & \\
 & A3C & POfD & \\
 &  & HAT, CHAT & \\
\end{tabular}
}
\caption[Comparison between different settings]{Comparison between different settings} \label{t:setting}
\end{table}

\section{Background}
\label{sec:bg}

\subsection{Reinforcement Learning and Deep Q Network}
\label{subsec:rl}
The standard reinforcement learning framework consists of an agent interacting with an environment which can be modeled as a Markov decision process (MDP). An MDP is defined by a tuple $M=\left<S, A, R, P, \gamma\right>$, where $S$ is the state space, $A$ the action space, $R: S \times A \rightarrow \R$ the reward function, $P(s'|s, a)$ the transition probability function, and $\gamma \in [0, 1)$ the discount factor. At each step, the agent observes a state $s \in S$ and takes an action $a \in A$ according to a policy $\pi$. The policy $\pi$ can be either deterministic, $\pi: S \rightarrow A$, or stochastic, $\pi: S \rightarrow P(A)$. On taking an action, the agent receives a reward $R(s, a)$ and reaches a new state $s'$ according to $P(s'|s, a)$. The goal of the agent is to find the policy $\pi$ which maximizes the discounted accumulative reward $\E_{\tau}\left[\sum_{t=0}^{\infty} \gamma^t R(s_t, a_t) \right]$, where $\tau$ denotes the trajectory obtained with $\pi$ and $P$.

For problems with discrete actions, the most popular approach nowadays is the Deep Q-network (DQN~\citep{DBLP:journals/nature/MnihKSRVBGRFOPB15}). The key idea of DQN is to learn an approximation of the optimal value function $Q^*$, which conforms to the Bellman optimality equation
\begin{equation*}
Q^*(s, a) = R(s, a) + \gamma \E_{s' \sim P(s'|s, a)} \left[\max_{a' \in A} Q^*(s', a')\right].
\end{equation*}
The optimal policy is then defined by $Q^*$ as $\pi(s) = \argmax_{a' \in A} Q^*(s, a')$.
The value-function is approximated by a neural network $Q(s,a;\theta)$ with parameter $\theta$ where the parameter is learned by minimizing the temporal difference loss:
\begin{align*}
Q_{target} &= r + \gamma \max_{a' \in A} Q(s', a';\theta^-) \\
L_{TD}(\theta) &= \E_{(s, a, r, s') \sim D} \left[(Q_{target} - Q(s, a;\theta))^2 \right],
\end{align*}
where $D$ is a distribution of transitions $(s, a, r=R(s, a), s' \sim P(s'|s, a))$ drawn from a replay buffer of previously observed transitions, and $\theta^-$ is the parameter of a separate target network which is copied from $\theta$ regularly to stabilize the target Q-values. Double Q-learning~\citep{DBLP:conf/aaai/HasseltGS16} is an enhancement of DQN where the target value is calculated by replacing $\max_{a' \in A} Q(s', a';\theta^-)$ with $Q(s', \argmax_{a' \in A} Q(s', a'; \theta); \theta^-)$. This modification reduces the overestimation of target value created with the original update rule.

\subsection{Deep Q-learning from Demonstration}
Deep Q-learning from Demonstration (DQfD~\citep{DBLP:conf/aaai/HesterVPLSPHQSO18}) is a state-of-the-art method to leverage demonstration data to accelerate the learning process of DQN. The agent is pre-trained on demonstration data to obtain better initial parameters before any interaction with the environment.  It keeps demonstration data in a prioritized replay buffer~\citep{Schaul2016} permanently, 
and give additional priority to demonstration data to increase the probability that they are sampled. DQfD applies a combination of four losses: the typical one-step double Q-learning loss ($L_{TD}$), N-step double Q-learning loss ($L_N$), supervised large margin classification loss ($L_E$), and L2 regularization loss. The overall loss is thus
\begin{equation*}
L(\theta) = L_{TD}(\theta) + \lambda_1 L_N(\theta) + \lambda_2 L_E(\theta) + \lambda_3 || \theta ||^2_2.
\end{equation*}
The typical one-step temporal difference loss and N-step temporal difference loss are used to obtain the optimal Q-value conforming to the Bellman equation, where the N-step loss is
\begin{align*}
L_{N}(\theta) &= \E_{(s, a, r, s') \sim D} \left[(Q_{N} - Q(s, a;\theta))^2 \right],\\
Q_N &= r_t + \gamma r_{t+1} + \gamma^2 r_{t+2} + ... + \gamma^{N-1} r_{t+N-1} + \max_a \gamma^N Q(s_{t+N}, a) .
\end{align*}
The large margin classification loss~\citep{DBLP:conf/pkdd/PiotGP14} is defined as
\begin{equation*}
L_E(\theta) = \max_{a \in A}\left[Q(s, a; \theta) + M\ind\left[a \neq a_E\right]\right] - Q(s, a_E),
\end{equation*}
where $a_E$ represents the action that the expert took in state $s$, $M$ is a positive constant value which means margin and $\ind[\cdot]$ is indicator function. L2 regularization loss is applied to parameters of the network to prevent overfitting on demonstration data. 
All losses are applied in both pre-training and reinforcement learning phases, whereas the supervised loss is only applied with demonstration data.

\subsection{Deep Exploration via Bootstrapped DQN}
Exploration is an important issue in RL. E.g., epsilon-greedy is commonly used but it does not exploit any information. Bootstrapped DQN~\citep{DBLP:conf/nips/OsbandBPR16} is a modification of DQN to improve exploration during training. In practice, the network is built with $K \in \N$ outputs, each representing a Q-value function estimation $Q_k(s, a; \theta)$. 
Each output head is trained against its own target network $Q_k(s, a; \theta^-)$ and 
is updated with its own bootstrapped data from the replay buffer. The parameters of each head are initialized independently, while the gradient of each update is normalized with $1/K$. During training, a single head is sampled at the beginning of each epoch, and the agent takes the optimal policy corresponding to the sampled Q-value approximation function for the duration of the episode. This allows the agent to conduct a more consistent exploration as compared to other common dithering strategies such as $\epsilon$-greedy. To keep track of the bootstrapped data for each head, we attach to each transition data in the replay buffer a boolean mask $m \in \{0, 1\}^K$ indicating which heads are privy to this data. The masks are drawn from an identical Bernoulli distribution independently ($m_i \sim Ber(p), \forall i \in 1 \ldots K$). However, their experiments show that the performance of bootstrapped DQN is not influenced by difference choices of $p$, and that all outperform the original DQN. Hence in practice, to increase computational efficiency, we simply share all the data between each head ($p=1$).

\subsection{Noisy Networks for Exploration}
NoisyNet~\citep{DBLP:journals/corr/FortunatoAPMOGM17} is an alternative approach to improve the efficiency of exploration in RL, where the parameters in the output layer of a network are perturbed by noise. The noisy parameter $\theta$ of $Q(s, a; \theta)$ is represented by $\theta = \mu + \Sigma \odot \varepsilon$, where $\zeta = (\mu, \Sigma)$ is a set of vectors of learnable parameters, $\varepsilon$ is a vector of zero-mean noise sampled from the standard normal distribution, and $\odot$ stands for element-wise multiplication. A noisy linear layer with $p$ inputs and $q$ outputs is then represented by
\begin{equation*}
y = (\mu_w + \sigma_w \odot \varepsilon_w)x + \mu_b + \sigma_b \odot \varepsilon_b,
\end{equation*}
where $\mu_w + \sigma_w \odot \varepsilon_w$ and $\mu_b + \sigma_b \odot \varepsilon_b$ replace the weight matrix and bias vector of typical linear regression. The parameters $\mu_w \in \R^{q \times p}, \mu_b \in \R^q, \sigma_w \ in \R^{q \times p}, \sigma_b \in \R^q$ are learnable parameters and $\varepsilon_w \in \R^{q \times p}, \varepsilon_b \in \R^q$ are noise variables.
The agent samples a new set of variables after each update step and follows the optimal policy corresponding to the sampled Q-value function. The noise of the online network, target network, and online network in double DQN are sampled independently. The loss function for noisy double DQN is defined as
\begin{align*}
&\bar{L}(\zeta) = \\
&\E_{\varepsilon, \varepsilon', \varepsilon''}\left[\E_{(s, a, r, s') \sim D}[r + \gamma Q(s', a^*, \varepsilon'; \zeta^-) - Q(s, a, \varepsilon; \zeta)]^2\right] \\
&a^* = \argmax_{a \in A} Q(s', a, \varepsilon''; \zeta).
\end{align*}

\section{Active Deep Q-learning with Demonstration}
\label{sec:pm}
In this section, we first describe a new problem setting, then propose an uncertainty-based query strategy to address the problem, after which we introduce two ways to estimate the uncertainty of a deep Q-network given an observed state with bootstrapped DQN and noisy DQN, separately.

\subsection{Problem Setup}
We proposed a new framework named Active Reinforcement Learning with Demonstration (ARLD) to improve the demonstration efficiency, which is not considered in previous RLED works. In ARLD, we consider the standard RL framework introduced in \ref{subsec:rl}. In addition, we assume there is an expert $\pi^+$ which performs well on the task we seek to learn. Notice that $\pi^+$  does not need to be optimal or deterministic, which is common for human experts. The agent interacts with the expert by asking what action to take only when it is not sure what to do given the observed state at each step. The algorithm then decides whether to take the action given by the expert, or to just take the action given by the agent's policy. The main challenge of ARLD is to decide when to query from the expert so that the agent can indeed benefit from the obtained demonstration. As with active learning, our goal is to improve RL by making as few queries as possible. More precisely, given a limited query budget, we seek to enable the agent to learn to solve the task in as few steps as possible. Below, we discuss uncertainty sampling~\citep{DBLP:journals/corr/LewisG94}, which one of the simplest and most commonly used query frameworks in active supervised learning.

\subsection{Query Strategy with Adaptive Uncertainty Threshold}
Uncertainty sampling is one of the simplest and most commonly used query frameworks in active learning~\citep{settles.tr09}. In this framework, an active learner estimates the uncertainty of a pool of unlabeled instances and submits queries for those it is least certain how to label. It is challenging to apply active learning with deep neural networks, as good deep models typically require large amounts of data. Recent work has shown that uncertainty can be estimated by taking advantage of specialized models such as Bayesian neural networks~\citep{DBLP:conf/icml/GalIG17}. However, to improve RL by requesting an expert demonstration, we require a online query strategy that takes advantage of uncertainty.

In our setting, at each step, before the agent takes an action, it decides whether to query the expert. A naive way to solve this problem is to make the decision with a fixed threshold: the agent asks expert for demonstration once the uncertainty of an observed state exceeds the threshold; otherwise it takes 
the action which maximizes the estimated Q-value.
However, it is difficult to find a proper threshold when the distribution of uncertainty 
keeps changing; 
the discrepancy between different tasks also makes this difficult. One proposal is to adjust the threshold with a fixed adjustment factor to work with the online Query by Committee (QBC)~\citep{DBLP:conf/ijcnn/KrawczykW17}, but it is still difficult to choose a good adjustment factor for all tasks, especially when the uncertainty measurement is unbounded and its magnitude unknown.

We propose an adaptive method which enables the agent to adjust its query policy during training time without any prior knowledge of the task. Each time the agent makes a decision, it compares the uncertainty of the current state with that estimated in recent steps. If the current state uncertainty is larger than a given proportion of recent steps, the agent queries the expert for demonstration; 
otherwise it determines its own action. 
In this algorithm, we decide whether to ask the expert given the parameters $N_r$ and $t_{query}$, representing the amount of recent steps we consider and the proportion of recent steps for which the state uncertainty must be higher than the current state uncertainty. In practice, we use a dequeue to maintain the uncertainty of recent steps and a balanced binary search tree (BST) to keep these uncertainties in order so that we can make the decision in $O(log_2N_r)$ complexity for each step. The pseudocode of the algorithm is provided in Algorithm~\ref{alg:qs}.

The performance of the algorithm depends on the choice of uncertainty estimation. Below, we propose two methods to estimate the uncertainty. One is based on bootstrapped DQN and the other one is based on noisy network.

\begin{algorithm}[ht]
\caption{Adaptive Query Strategy with Uncertainty}
\label{alg:qs}
\begin{algorithmic}[1]
\Require uncertainty $U_t$, reference size $N_r$, proportion threshold $t_{query} \in [0, 1]$, recent uncertainty dequeue $D$, recent uncertainty BST $B$
\Ensure asking $\in [\mathit{TRUE}, \mathit{FALSE}]$, $D$, $B$
\State $idx \gets$ size of $D \times t_{query}$
\State $U_{threshold} \gets B[idx]$
\If{$U_t > U_{threshold}$}
	\State asking $\gets \mathit{TRUE}$
\Else
	\State asking $\gets \mathit{FALSE}$
\EndIf
\If{size of D $\geq R_r$}
	\State $U_{del} \gets$ $D.pop\_left()$
    \State remove $U_{del}$ from $B$
\EndIf
\State add $U_t$ into $D$, $B$
\end{algorithmic}
\end{algorithm}

\subsection{Divergence of Bootstrapped DQN}
Bootstrapping is a commonly used technique in statistics to estimate a sampling distribution. In bootstrapped DQN~\citep{DBLP:conf/nips/OsbandBPR16}, multiple value function heads $Q_k(s, a;\theta)$ are used to approximate a distribution over Q-values. There are several ways to estimate uncertainty with these bootstrapped heads, including calculating the entropy of the voting distribution or averaging the variance of action values predicted by each head. In this work, we consider each head as a distribution and estimate the uncertainty using Jensen-Shannon divergence, which is a well-known method to measure the similarities between multiple distributions.

Notice that while the agents in bootstrapped DQN tend to behave differently at less critical states because the Q-values of each action might be close to each other, the JS divergence in this situation will be low since the distributions of Q-values are similar. Thus, we can prevent from asking these unimportant states by estimating JS divergence as uncertainty. For example, when considering an environment with two actions and two bootstrapped value functions, if the two heads predict (1, 0) and (0, 1) respectively, intuition dictates that the state should be more uncertain than (0.5, 0.4) and (0.4, 0.5) in another state.  
However, with the voting method, both states are considered equally uncertain, as the two heads vote for different actions in both cases; JS divergence, by contrast, distinguishes the difference between them. Another advantage of JS divergence is that as it is a bounded function, its value is more meaningful and 
easier to use as a threshold in different environments.

To measure the JS divergence between the bootstrapped heads, we first normalize the Q-values and actions using softmax to obtain a policy distribution. For each head $Q_k(s, a;\theta$),
$$ \pi_k(a | s;\theta) = e^{Q_k(s, a;\theta)} / \sum_{a'} e^{Q_k(s, a';\theta)}.$$
Given this policy distribution, we estimate the uncertainty by calculating the Jensen-Shannon divergence of the policy distribution between each head, yielding
$$ U_D = JS(\pi_1, \pi_2, ..., \pi_K) = H(\frac{1}{K}\sum_k \pi_k) - \frac{1}{K}\sum_k H(\pi_k),$$
where $H(\pi)$ is the Shannon entropy of distribution $\pi$ and $K$ the number of bootstrapped heads.

\subsection{Predictive Variance of Noisy DQN}
For our second estimate of uncertainty, we evaluate the predictive variance of a noisy network. Previous work has shown the effectiveness of estimating uncertainty by the predictive variance of a Bayesian convolution network in classification active learning~\citep{DBLP:conf/icml/GalIG17}. Other works have shown that injecting noise into the parameter space improves the exploration process in deep reinforcement 
learning~\citep{DBLP:journals/corr/FortunatoAPMOGM17,DBLP:journals/corr/PlappertHDSCCAA17}. 
Combining these two ideas, we use the noisy network as an exploration policy and estimate uncertainty using its predictive variance. We replace the fully connected layer in the output layer with a noisy fully connected layer. The corresponding noisy output layer can be seen as Bayesian linear regression:
\begin{gather*}
Q(s, a;\theta) = w_a\phi(s) + b_a,\\
w_a \sim N(\mu^{(w_a)}, \Sigma), \Sigma = diag((\sigma^{(w_a)})^2)\\
b_a \sim N(\mu^{(b_a)}, (\sigma^{(b_a)})^2),
\end{gather*}
where $\phi(s) \in \R^p$ is input to the last layer. $w_a \in \R^p, b_a \in \R$ represents the variables corresponding to action $a$, and $\mu^{(w_a)}$, $\sigma^{(w_a)}$, $\mu^{(b_a)}$, and $\sigma^{(b_a)}$ are the parameters actually learned by the model, representing the mean and noise level of $w_a$ and $b_a$ respectively.

Given the posterior distribution of the parameters, we derive the predictive variance as
\begin{align*}
Var[Q(s, a)] &= Var[w_a\phi(s) + b_a] \\
             &= Var[w_a\phi(s)] + Var[b_a] \\
             &= \phi(s)^T\Sigma\phi(s) + (\sigma^{(b_a)})^2 .
\end{align*}

The variance of each action measures the lack of confidence with respect to this action. We take the variance of the action with the largest Q-value as our uncertainty:
\begin{equation*}
U_V = Var(Q(s, \argmax_a Q(s, a))),
\end{equation*}
which translates to the lack of confidence for the action the agent would take at the step. By querying the states with low confidence, we avoid bad moves leading to task failure and explicitly teach the agent which is the proper action of the state.

\section{Experiments}
\label{sec:exp}
In this section, we describe the environments used for the evaluation as well as the experimental setup. To focus on the effectiveness of each query strategy, we show the experimental results of methods based on the bootstrapped and noisy network separately, after which we present results evaluating the effect of the query proportion threshold of the proposed method.

\begin{figure*}[tb]
	\begin{minipage}[t]{\textwidth}
        \centering
        \includegraphics[width=\textwidth]{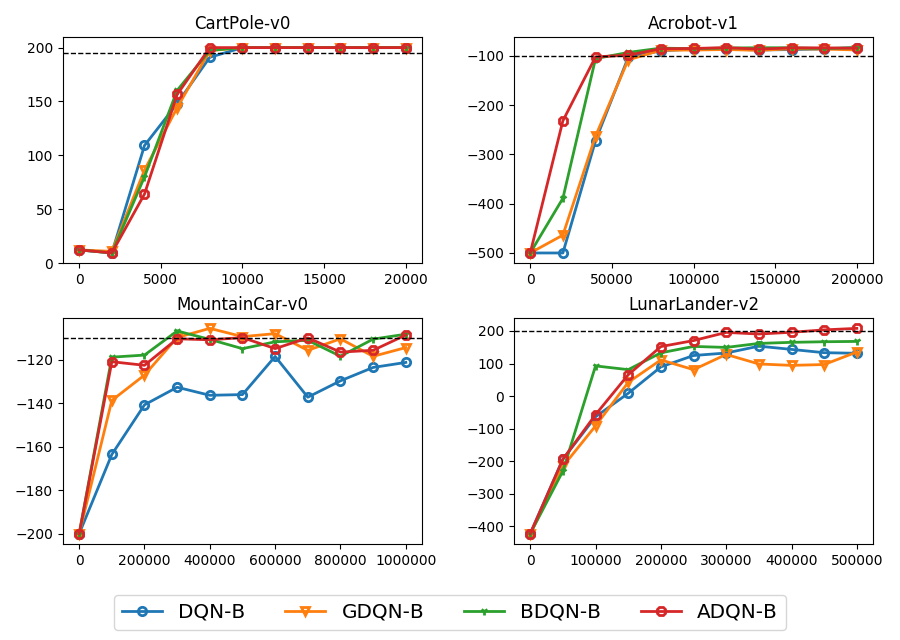}
        \subcaption{Comparison between methods with bootstrapped DQN}
        \label{fig:compare_bootstrap}
    \end{minipage}
    \begin{minipage}[t]{\textwidth}
        \centering
        \includegraphics[width=\textwidth]{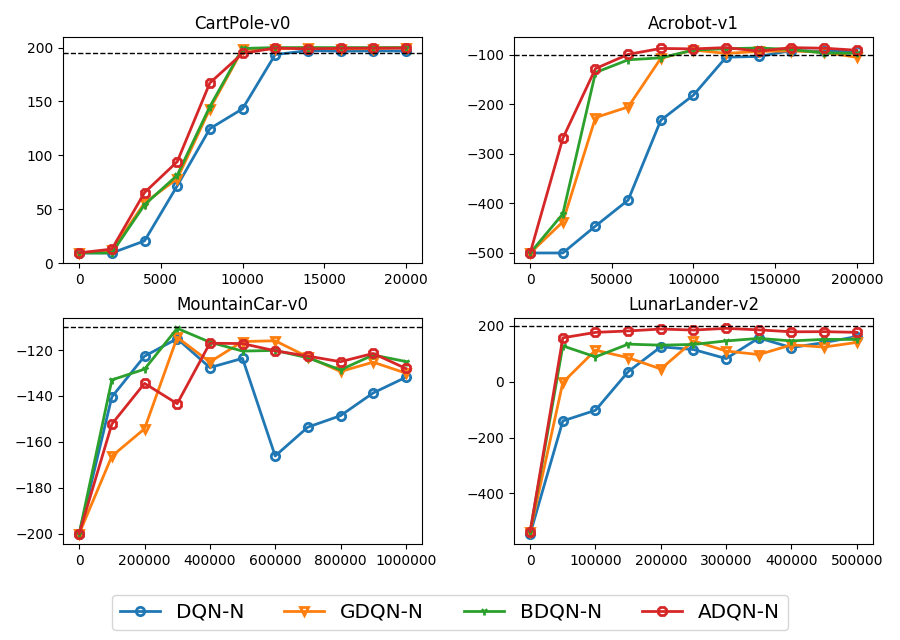}
        \subcaption{Comparison between methods with bootstrapped DQN}
        \label{fig:compare_noisy}
    \end{minipage}
    \caption{Comparison between different ARLD methods. 
    The dashed lines indicate the score of solving the task.}
    \label{fig:compare_adqn}
\end{figure*}

\subsection{Experimental Setup}
\label{subsec:exp_setup}
We use four different environments for evaluation: (1) Cart-Pole, (2) Acrobot, (3) Mountain Car, and (4) Lunar Lander, all of which are included in OpenAI Gym~\citep{DBLP:journals/corr/BrockmanCPSSTZ16}. Among them Cart-Pole is the simplest task and Lunar Lander is the most complicated one. The target score to mark each task as solved is listed in Table~\ref{t:expert_stat}.

For each environment, we evaluated six different methods based on bootstrapped~\citep{DBLP:conf/nips/OsbandBPR16} or noisy~\citep{DBLP:journals/corr/FortunatoAPMOGM17} networks separately. The methods are:
\begin{enumerate}
	\item \textbf{DQN}: Prioritized Double DQN trained without any demonstration
	\item \textbf{DQfD}: Deep Q-learning from Demonstration~\citep{DBLP:conf/aaai/HesterVPLSPHQSO18}
	\item \textbf{GDQN}: Greedy query strategy which queries all states until budget is all spent
	\item \textbf{BDQN}: Bernoulli query strategy, queries states at fixed probability
	\item \textbf{ADQN}: Active DQN, queries states according to proposed query strategy and uncertainty estimation
	\item \textbf{ADQNP}: Active DQN with DQfD pretraining.
\end{enumerate}
The key differences between the methods are summarized in Table~\ref{t:methods}.

\begin{table}[ht]
\resizebox{\linewidth}{!}{
\begin{tabular}{lccccc}
\hline
       & Demonstration & Pre-training & Interaction & Query criterion \\
\hline \hline
DQN  & no & no & no & no \\ 
\hline
DQfD & yes & yes & no & no \\
\hline
GDQN & yes & no & yes &  greedy \\
\hline
BDQN & yes & no & yes & Bernoulli \\
\hline
ADQN & yes & no & yes & uncertainty \\
\hline
ADQNP & yes & yes & yes & uncertainty \\
\hline \hline
\end{tabular}
}
\caption{Comparison between methods} \label{t:methods}
\end{table}

We tuned the basic parameters such as the learning rate and discount factor for DQN on each environment to ensure reasonable learning progress and then fixed these parameters for all six methods, as listed in Table~\ref{t:task_params}. The network structure applied in all environments was identical: two fully connected hidden layers with 64 neurons followed by another fully connected layer to the Q-Values for each action. The layers all used rectified linear units (ReLU) for non-linearity. We trained the networks using Adam and a $\epsilon$-greedy policy with $\epsilon$ annealed linearly from 0.9 to 0.01. We set the parameters of prioritized replay according to~\citep{Schaul2016}. For bootstrapped DQN, we used 10 bootstrap heads with normalized gradient and shared all the data as~\citep{DBLP:conf/nips/OsbandBPR16}. For the noisy networks, we used factorized noise and followed the initialization and  hyperparameter values from \citep{DBLP:journals/corr/FortunatoAPMOGM17}.

For DQfD, we did not use L2 regularization loss or N-step temporal difference loss, as they brought no benefit to training in our experiments. We set the expert margin $M = 0.8$ as~\citep{DBLP:conf/aaai/HesterVPLSPHQSO18} and tuned the supervised loss weight $\lambda$ in $\{10^{-5}, 10^{-4}, 10^{-3}, \ldots , 1\}$. The number of demonstration data and pre-training steps were set to allow DQfD learn a better initial policy than learning from a scratch.

For each query, all ARLD methods receives five consecutive expert demonstrations until the end of the episode. The query threshold $t_{query}$ is tuned in $\{0.05, 0.1, 0.3, 0.5\}$. For ADQNP, the number of demonstration for pretraining is half of DQfD and the query budget is half of ADQN, resulting in the same number of total demonstrations. All task-specific parameters are all listed in Table~\ref{t:task_params}.

\begin{table}[ht]
\resizebox{\linewidth}{!}{
\begin{tabular}{lcccc}
\hline
& Mean score/std & Min. score & Avg. steps & Target score\\
\hline \hline
Cart-Pole    & 166.77$\pm$39.14 & 93 & 166.77 & 195 \\
\hline
Acrobot      & -128.25$\pm$66.86 & -489 & 128.25 & -100 \\
\hline
Mountain Car & -134.0$\pm$27.52 & -158 & 134 & -110 \\
\hline
Lunar Lander & 155.18$\pm$55.58 & -16.63 & 784.92 & 200 \\
\hline \hline
\end{tabular}
}
\caption[Expert statistics]{Expert statistics} \label{t:expert_stat}
\end{table}

\begin{table}[ht]
\resizebox{\linewidth}{!}{
\begin{tabular}{l|c|c|c|c}
\hline
                   & Cart-Pole & Acrobot & Mountain Car & Lunar Lander \\
\hline \hline
Discount factor    & 0.9       & 0.99    & 0.99         & 0.99         \\
Learning rate      & 0.0001    & 0.0001  & 0.001        & 0.001        \\
Training steps     & 20000     & 200000  & 500000       & 500000       \\
\# of demo/budget  & 200       & 100     & 500          & 3000         \\
Memory size        & 10000     & 100000  & 100000       & 100000       \\
Pre-training steps & 10000     & 10000   & 10000        & 30000        \\
DQfD $\lambda$     & 0.00001   & 1       & 1            & 1            \\
ADQN-B $t_{query}$ & 0.05      & 0.3     & 0.1          & 0.3  	       \\
ADQN-N $t_{query}$ & 0.5       & 0.3     & 0.3          & 0.1          \\
\hline \hline
\end{tabular}
}
\caption[Task-specific parameters]{Task specific parameters} \label{t:task_params}
\end{table}

To obtain an expert for each environment, we saved the prioritized double DQN models during training and evaluated each over 100 episodes. Then we chose as the environment expert a model that (a) did not perfectly solve the task (b) had reached low performance variance (c) still solved the task before the end of the episode. The choice ensures the experts to be realistic rather than idealistic. These experts were used to collect demonstration data in DQfD and perform interactive demonstration in ADQN. The evaluation statistics of these experts are shown in Table~\ref{t:expert_stat}. In section~\ref{subsec:expert_quality}, we show the effect of different artificial expert settings on ADQN and DQfD.

All of the experiments were repeated 20 times with different random seeds. The figures show the median of the results over 20 trials. The y-axis indicates the averaged test score, where the test scores in each trial were computed at a fixed frequency by executing 20 test episodes without exploration.

\begin{figure*}[htb]
	\begin{minipage}[t]{\textwidth}
        \centering
        \includegraphics[width=\textwidth]{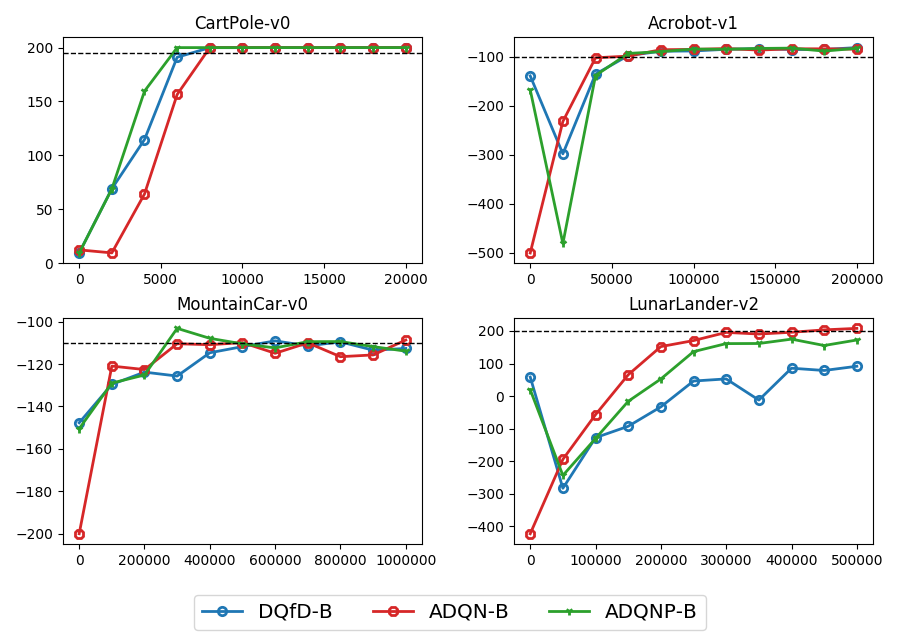}
        \subcaption{ADQN, DQfD and ADQNP based on bootstrapped DQN}
        \label{fig:dqfd_b}
    \end{minipage}
    \begin{minipage}[t]{\textwidth}
        \centering
        \includegraphics[width=\textwidth]{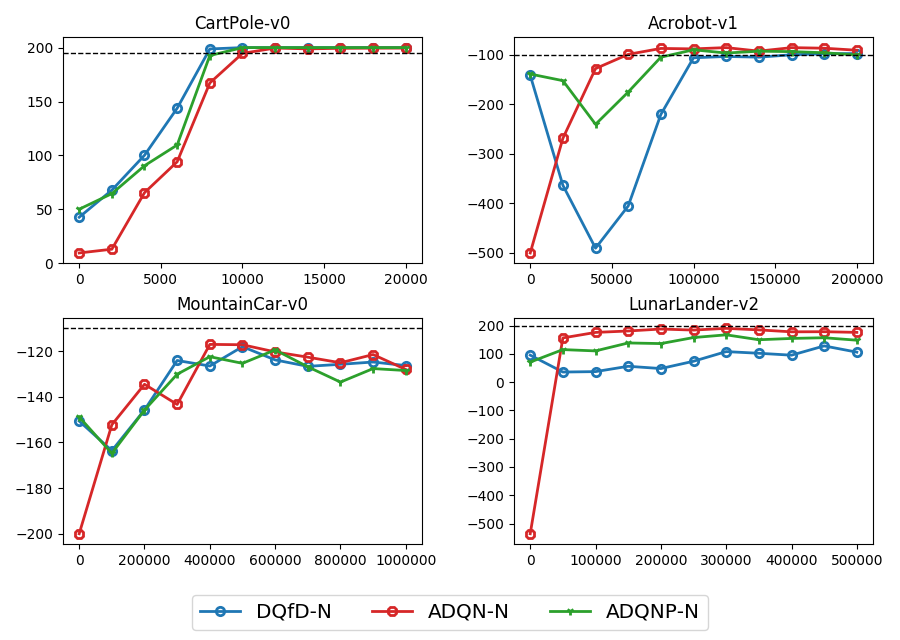}
        \subcaption{ADQN, DQfD and ADQNP based on noisy DQN}
        \label{fig:dqfd_n}
    \end{minipage}
    \caption{Comparison between ADQN, DQfD and ADQNP. 
    The dashed lines indicate the score of solving the task.
    }
    \label{fig:compare_dqfd}
\end{figure*}

\begin{figure*}[htb]
	\begin{minipage}[t]{\textwidth}
        \centering
        \includegraphics[width=\textwidth]{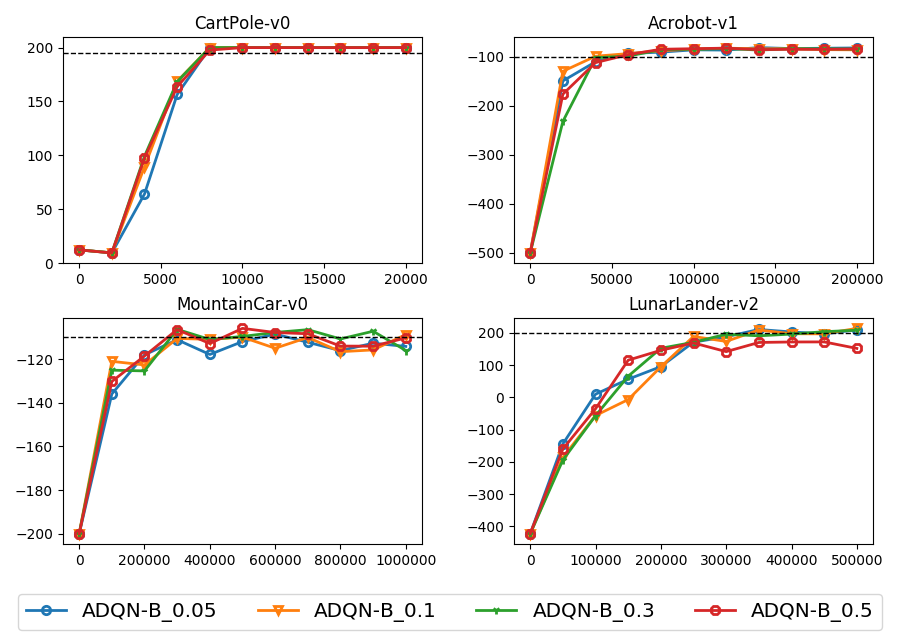}
        \subcaption{Active DQN based on bootstrapped DQN}
        \label{fig:mc_tq_n}
    \end{minipage}
    \begin{minipage}[t]{\textwidth}
        \centering
        \includegraphics[width=\textwidth]{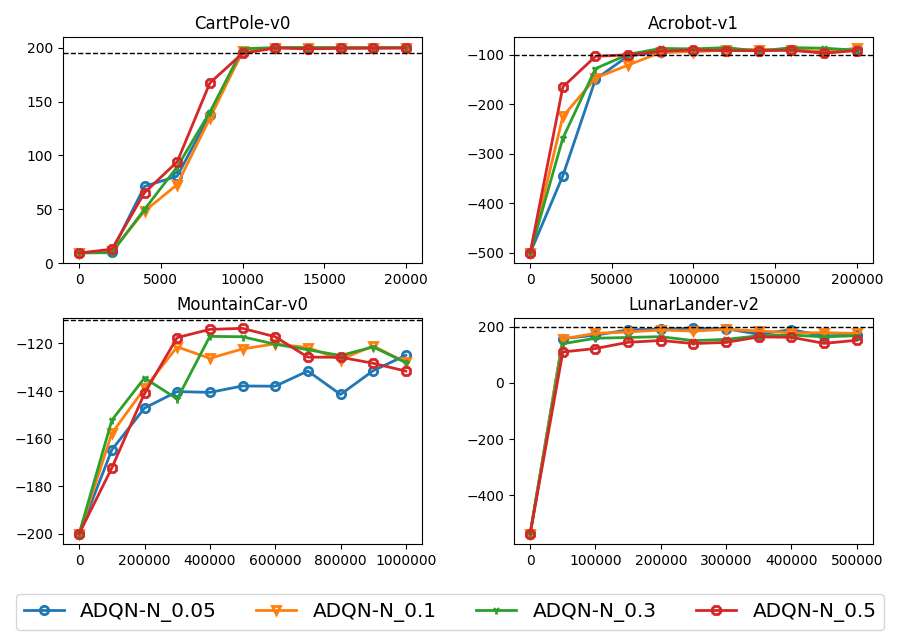}
        \subcaption{Active DQN based on noisy DQN}
        \label{fig:lunar_tq_n}
    \end{minipage}
    \caption{Comparison between different query thresholds among \{0.05, 0.1, 0.3, 0.5\}. 
    }
    \label{fig:compare_tq}
\end{figure*}

\subsection{Comparison between ARLD methods}
We first compare the methods without pretraining, i.e., DQN, GDQN, BDQN and ADQN. Table~\ref{t:step_solve} lists the median of number of steps that each method takes to solve the tasks in 20 trials. Among methods based on bootstrapped DQN, the proposed ADQN outperforms other methods in all environments and yields significant improvements in Acrobot, Mountain Car, and Lunar Lander. For methods based on noisy network, ADQN also achieves best performance in three out of four tasks, and improves the learning progress in Acrobot and Lunar Lander dramatically. The strength of ADQN over DQN again confirms the usefulness of interacting with demonstration. Most importantly, the advantage of ADQN over GDQN and BDQN validates that ADQN allows a more effective use of the demonstration efforts by querying strategically at the important moments.

Fig.~\ref{fig:compare_adqn} shows the learning curves of ARLD methods based on the bootstrapped or noisy network separately. The results demonstrates that the methods with demonstration not only outperform the original DQN in general, but also achieve higher score than the realistically-simulated experts. Among the methods with demonstration, ADQN is often the most competitive one, especially in the hardest task of Lunar Lander, which was solved by ADQN with fewer steps and a higher score.


\begin{table}[ht]
\resizebox{\linewidth}{!}{
\begin{tabular}{l|c|c|c|c}
\hline
       & Cart-Pole & Acrobot    & Mountain Car & Lunar Lander \\
\hline \hline
DQN-B  & 8000      & 57000      & 210000       & 260000       \\
GDQN-B & 8000      & 45000      & 170000       & 310000       \\
BDQN-B & 7500      & 31000      & 100000       & 217500       \\
ADQN-B & 7000 & \bf{25000} & \bf{85000}   & \bf{205000}  \\
\hline
DQfD-B & 7500      & 38000      & 140000       & 500000       \\
ADQNP-B & \bf{6000} & 49000     & 135000       & 267500       \\
\hline \hline
DQN-N  & 13000     & 114000     & 190000       & 355000       \\
GDQN-N & 10000     & 73000      & 295000       & 500000       \\
BDQN-N & 10000     & 55000      & \bf{170000}  & 160000       \\
ADQN-N & 9500      & \bf{8000}  & 230000       & \bf{47500}   \\
\hline
DQfD-N & \bf{8000} & 95000      & 300000       & 402500       \\
ADQNP-N & \bf{8000} & 75000     & 255000       & 212500       \\
\hline \hline
\end{tabular}
}
\caption{Median number of step to solve the task. The bold numbers indicate the best performance on that task.}
\label{t:step_solve}
\end{table}

\subsection{Comparison with Pretraining Methods}
Next, we compare DQfD with ADQN to understand the effect of collecting expert demonstration \textit{before} or \textit{during} training. We also design ADQNP as a simple mixture between the two. The results in Table~\ref{t:step_solve} show that ADQN usually solves the tasks with fewer steps than DQfD or ADQNP, except for the simplest task of Cart Pole. ADQNP also often improves over DQfD when the tasks gets harder. The results justify the effectiveness of leveraging expert demonstration \textit{during} training. 

For simulating real-world scenario where the demonstrating humans may not always be perfect, our simulated experts are designed to be realistic but imperfect. Then, methods with pretraining need to take additional steps in the beginning to correct the policy learned from the imperfect experts. This situation explains the performance dropping in the beginning of DQfD for Acrobot and Lunar Lander, as shown in Fig~\ref{fig:compare_dqfd}. ADQN, on the other hand, demonstrates better ability to leverage the imperfect demonstrations to aid RL.

\subsection{Effect of Query Proportion Threshold}
Active DQN uses two parameters: the number of recent steps for which we compare the uncertainty ($N_r$) and the proportion threshold that determines whether to make a query given recent steps ($t_{query}$). Since the uncertainty distribution usually changes smoothly, the value of $N_r$ effects the performance little compared to parameter $t_{query}$. In Fig.~\ref{fig:compare_tq} we plot the performance given different values of $t_{query}$: we observe that in most cases, different choices of $t_{query}$ perform similarly. Moreover, in Table~\ref{t:task_params} we see that the best values of $t_{query}$ are either 0.1 or 0.3 in Acrobot, Mountain Car, and Lunar Lander; in Cart-Pole, the only exception, the result shows the least variance between choices of $t_{query}$. Thus, the performance of Active DQN is not sensitive to the parameter $t_{query}$, and it is easy to choose a value between 0.1 and 0.3 that optimizes performance for all kinds of tasks.

\begin{figure*}[htb]
	\begin{minipage}[t]{\textwidth}
        \centering
        \includegraphics[width=\textwidth]{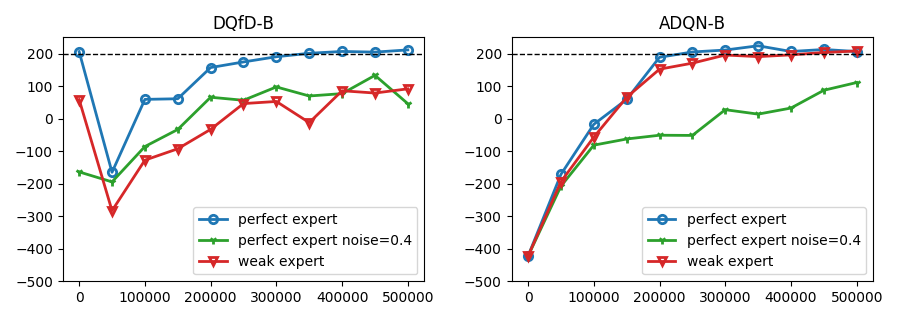}
        \subcaption{Effect of expert's quality on DQfD-B and ADQN-B}
        \label{fig:expert_b}
    \end{minipage}
    \begin{minipage}[t]{\textwidth}
        \centering
        \includegraphics[width=\textwidth]{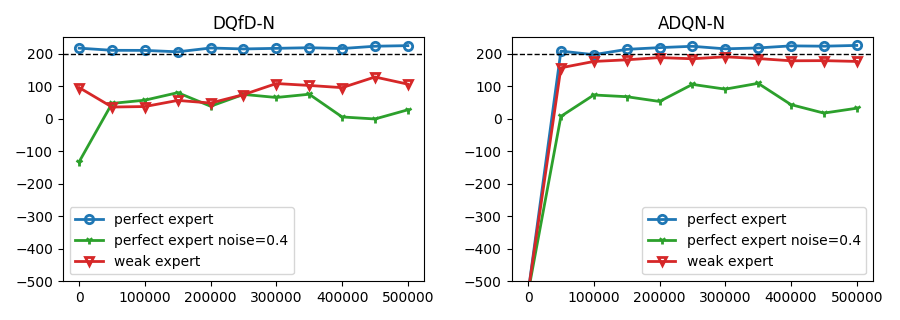}
        \subcaption{Effect of expert's quality on DQfD-N and ADQN-N}
        \label{fig:dqfd_n}
    \end{minipage}
    \caption{Comparison between perfect expert, prefect expert with 40\% random action, and weak expert.
    }
    \label{fig:experts}
\end{figure*}

\subsection{Effect of Expert's Quality on ADQN and DQfD}
\label{subsec:expert_quality}
In this section, we compare three different approaches to obtain the expert agents used in the experiment. First of all, the perfect experts are DQN agents trained on each task until convergence. These well-trained experts can solve their tasks perfectly and efficiently. Second, we obtain weaker experts by applying random noise to the perfect expert. That is, each time an expert is going to make a demonstration, there is a probability the expert will do a random action rather than following the perfect expert's policy. The random behaviors sometimes lead to the end of an episode directly, therefore even though the expert is able to make optimal choices at most of the time, it still might fail to solve the task at the beginning of an episode. Last, as described in section~\ref{subsec:exp_setup} and \ref{t:expert_stat}, we saved the temporary models in the process of training a DQN agent and selected one of them to be a weak expert. Compared to the noisily-acting weak experts, these policy-consistent weak experts act more consistently through an episode, hence their behavior are more similar to a human expert.

Figure~\ref{fig:experts} demonstrates the effect of expert's quality on ADQN and DQfD in Lunar Lander. We experiment with both noisy and bootstrapped network structures along with the three experts mentioned above. The figure shows both DQfD and ADQN can learn efficiently and solve the task within few steps with an perfect expert. On the contrary, both of their performances suffer from the noisily-weak expert. They not only learned slower than learning with perfect experts, but also converge at a worse score. However, while working with the policy-consistent weak expert, though DQfD still perform poorly, ADQN converge at higher scores which are close to ones achieved by working with perfect experts. As a result, we found that ADQN is able to take the advantage of policy-consistent weak experts, which are similar to human experts.
\section{Conclusion and Future Work}
\label{sec:con}

In this work, we propose Active DQN, which improves RL with demonstration more efficiently with regard to human effort. We use DQfD to leverage demonstration data and propose a novel uncertainty-based query strategy which applies to diverse tasks. We provide two measurements of the uncertainty: the divergence of Bootstrapped DQN, and the predictive variance of Noisy DQN. Experimental results show that both of the proposed methods yield better performance than the expert and learn faster with the same number of demonstrations in different tasks.

As an initial work on Active Reinforcement Learning with Demonstration, the proposed method has achieved promising performance. A possible extension of this work is to apply it on RL algorithms such as DDPG which work on a continuous action space. A more difficult challenge is to work with off-policy methods such as policy gradient which require a different way to learn with demonstration.

%
%

\clearpage

\section*{Acknowledgement}
MS was supported by KAKENHI 17H00757.


\bibliographystyle{spbasic}      
\bibliography{reference}   

%
%

\end{document}